\pdfoutput=1

\documentclass[11pt]{article}

\usepackage{acl}

\usepackage{times}
\usepackage{latexsym}

\usepackage[T1]{fontenc}

\usepackage[utf8]{inputenc}

\usepackage{microtype}
\usepackage{hyperref}

\usepackage{booktabs}
\usepackage{graphicx}

\usepackage{xspace,mfirstuc,tabulary}

\usepackage{amsmath}
\usepackage{multirow}
\usepackage{array}

\newcommand{\bftab}{\fontseries{b}\selectfont}
\newcommand{\amazonreviews}{{azn}\xspace}
\newcommand{\gnad}{gnad\xspace}
\newcommand{\agnews}{{AG-news}\xspace}
\newcommand{\headqa}{hqa\xspace}
\newcommand{\hatespeech}{hate\xspace}
\newcommand{\solid}{solid\xspace}
\newcommand{\fasl}{\textsc{FASL}\xspace}

\newcommand{\rmetric}{$\mathcal{R}$\xspace}
\newcommand{\pmetric}{$\mathcal{P}$\xspace}
\newcommand{\fmetric}{$\mathcal{F}_1$\xspace}

\begin{document}
\title{Active Few-Shot Learning with \fasl}

\author{
Thomas M{\"u}ller$^1$,
Guillermo Pérez-Torró$^1$,
Angelo Basile$^{1,2}$ \and
Marc Franco-Salvador$^1$ \\
\vspace{-6pt}\\
$^1$ Symanto Research, Valencia, Spain\phantom{$^1$ } \\
\url{https://www.symanto.com}\\
\texttt{\{thomas.mueller,guillermo.perez,angelo.basile,marc.franco\}@symanto.com}
\vspace{6pt}\\
$^2$ PRHLT Research Center, Universitat Politècnica de València, Spain\phantom{$^2$ }\\
}

\maketitle              
\begin{abstract}
Recent advances in natural language processing (NLP) have led to strong text classification models for many tasks.
However, still often thousands of examples are needed to train models with good quality.
This makes it challenging to quickly develop and deploy new models for real world problems and business needs.
Few-shot learning and active learning are two lines of research, aimed at tackling this problem.
In this work, we combine both lines into \fasl, a platform that allows training text classification models using an iterative and fast process.
We investigate which active learning methods work best in our few-shot setup.
Additionally, we develop a model to predict when to stop annotating.
This is relevant as in a few-shot setup we do not have access to a large validation set.

\end{abstract}

\section{Introduction}

In recent years, deep learning has lead to large improvements on many text classifications tasks.
Unfortunately, these models often need thousands of training examples to achieve the quality required for real world applications.
Two lines of research aim at reducing the number of instances required to train such models: Few-shot learning and active learning.

Few-shot learning (FSL) is the problem of learning classifiers with only few training examples.
Recently, models based on natural language inference (NLI) \cite{bowman-etal-2015-large} have been proposed as a strong backbone for this task \cite{yin-etal-2019-benchmarking,yin-etal-2020-universal,halder-etal-2020-task,Wang2021EntailmentAF}.
The idea is to use an NLI model to predict whether a textual premise (input text) entails a textual hypothesis (label description) in a logical sense. 
For instance, ``\emph{I am fully satisfied and would recommend this product to others}'' implies ``\emph{This is a good product}''.
NLI models usually rely on cross-attention which makes them slow at inference time and fine-tuning them often involves updating hundreds of millions of parameters or more.
Label tuning (LT) \cite{labeltuning} addresses these shortcomings using Siamese Networks trained on NLI datasets to embed the input text and label description into a common vector space.
Tuning only the label embeddings yields a competitive and scalable FSL mechanism as the Siamese Network encoder can be shared among different tasks.

Active learning (AL) \cite{settles2009active} on the other hand attempts to reduce the data needs of a model by iteratively selecting the most useful instances.
We discuss a number of AL methods in S. \ref{sec:al_methods}.
Traditional AL starts by training an initial model on a seed set of randomly selected instances.
In most related work, this seed set is composed of at least 1,000 labeled instances.
This sparks the question whether the standard AL methods work in a few-shot setting.

A critical question in FSL is when to stop adding more annotated examples.
The user usually does not have access to a large validation set which makes it hard to estimate the current model performance.
To aid the user we propose to estimate the normalized test F1 on unseen test data.
We use a random forest regressor (RFR) \cite{ho1995random} that provides a performance estimate even when no test data is available.

\fasl is a platform for active few-shot learning that integrates these ideas.
It implements LT as an efficient FSL model together with various AL methods and the RFR as a way to monitor model quality.
We also integrate a user interface (UI) that eases the interaction between human annotator and model and
makes \fasl accessible to non-experts.

We run a large study on AL methods for FSL, where we evaluate a range of AL methods on
6 different text classification datasets in 3 different languages.
We find that AL methods do not yield strong improvements over a random baseline when applied to datasets with balanced label distributions.
However, experiments on modified datasets with a skewed label distributions as well as naturally unbalanced datasets show the value of AL methods such as \emph{margin sampling} \cite{lewis1994}.
Additionally, we look into performance prediction and find that a RFR outperforms stopping after a fixed number of steps.
Finally, we integrate all these steps into a single uniform platform: \fasl.

\section{Methods}

\subsection{Active Learning Methods}
\label{sec:al_methods}

\emph{Uncertainty sampling} \cite{lewis1994} is a framework where the most informative instances are selected using a measure of uncertainty.
Here we review some approaches extensively used in the literature \cite{settles2009active}.
We include it in three variants, where we select the instance that maximizes the corresponding expression:

\begin{itemize}
\item \textbf{Least Confidence}: With $\hat{y}$ as the most probable class: $- \mathrm{P} (\hat{y} \mid x)$
\item \textbf{Margin}: With $\hat{y_i}$ as ith most probable class:\\ $- \left[ \mathrm{P}( \hat{y_1} \mid x) - \mathrm{P}(\hat{y_2} \mid x) \right]$
\item \textbf{Entropy}:\\ $\mathrm{H}(Y) = - \sum_{j} \mathrm{P}( y_j \mid x) \log \mathrm{P}( y_j \mid x)$
\end{itemize}

Where $\mathrm{P}(Y \mid X)$ denotes the model posterior.
While \textit{uncertainty sampling} depends on the model output, \textit{diversity sampling} relies on the input representation space \cite{DASGUPTA20111767}.
\citet{Nguyen2004ActiveLU} cluster the instances and use the cluster centroids as a heterogeneous instance sample.
We experiment with three different methods: K-medoids \cite{Park2009ASA}, K-means \cite{1056489} and agglomerative single-link clustering (AC).
K-medoids directly finds cluster centroids which are real data points. For the others, we select the instance closest to the centroid.

Other research lines have explored different ways of combining both approaches. 
We adapt a two steps process used in computer vision \cite{DBLP:journals/corr/abs-1711-10856}.
First, we cluster the embedded instances and then sample from each group the instance that maximizes one of the uncertainty measures presented above.
We also implemented contrastive active learning (CAL) \cite{margatina-etal-2021-active}.
CAL starts with a small labeled seed set and finds k labeled neighbors for each data point in the unlabeled pool.
It then selects the examples with the highest average Kullback-Leibler divergence w.r.t their neighborhood.

\subsection{Few-shot Learning Models}
\label{sec:al_models}

We experiment with two FSL models.
The first model uses the zero-shot approach for Siamese networks and label tuning (LT) \cite{labeltuning}.
We encode the input text and a label description -- a text representing the label -- into a common vector space using a pre-trained text embedding model.

In this work, we use Sentence Transformers \cite{reimers-2019-sentence-bert} and, in particular, \textit{paraphrase-multilingual-mpnet-base-v2}\footnote{\url{https://tinyurl.com/pp-ml-mpnet}} a multilingual model based on \textit{roberta XLM} \cite{DBLP:journals/corr/abs-1907-11692}.
The dot-product is then used to compute a score between input and label embeddings.
LT consists in fine-tuning only the label embeddings using a cross-entropy objective applied to the similarity matrix of training examples and labels.
This approach has three major advantages: (i) training is fast as the text embeddings can be pre-computed and thus the model just needs to be run once.
As only the label embeddings are fine-tuned (ii) the resulting model is small and (iii) the text encoder can be shared between many different tasks.
This allows for fast training and scalable deployment.

The second model is a simple logistic regression (LR) model. We use the implementation of \textit{scikit-learn} \cite{scikit-learn}.
As feature representation of the text instances we use the same text embeddings as above.
In the zero-shot case, we train exclusively on the label description.

\begin{figure*}[t]
\centering
\resizebox{0.8\textwidth}{!}{%
\includegraphics[]{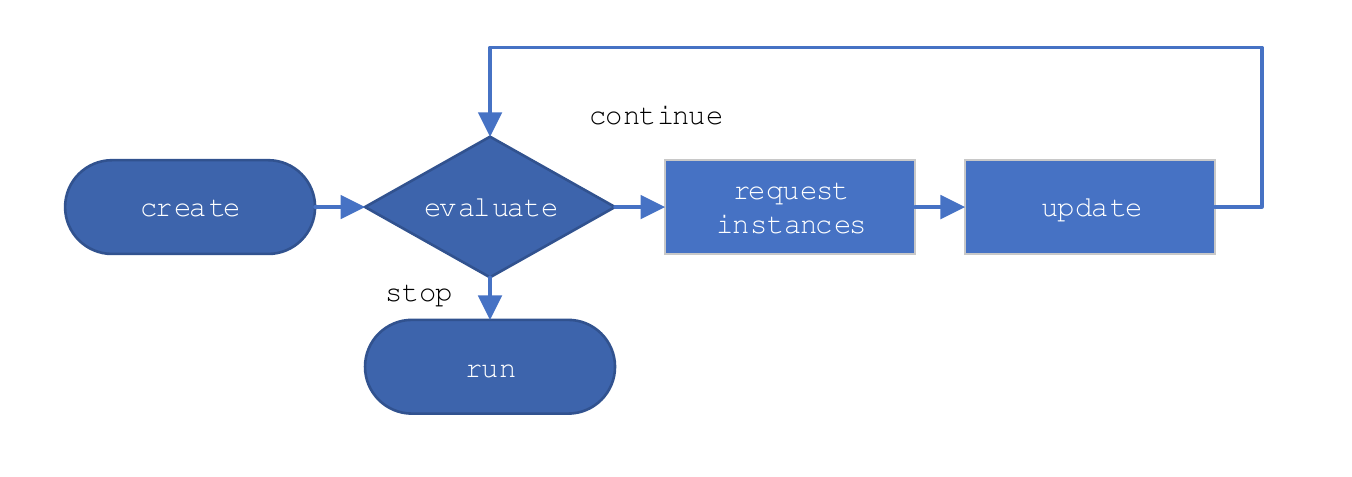}
}
\caption{\fasl API diagram.}
\label{fig:system_diagram}
\end{figure*}

\subsection{System Design and User Interface}
\label{sec:al_platform}

We implement the AL platform as a REST API 
with a web frontend.
Figure \ref{fig:system_diagram} shows the API flow.
The Appendix (S. \ref{sec:appendix}) contains screenshots of the UI.

The user first creates a new model (\verb!create!) by selecting the dataset and label set and label descriptions to use as well as the model type.
They can then upload a collection of labeled examples that are used to train the initial model.
If no examples are provided the initial model is a zero-shot model.

Afterwards the user can either request instances (\verb!request instances!) for labeling or run model inference as discussed below.
Labeling requires selecting one of the implemented AL methods. 
Internally the platform annotates the entire unlabeled dataset with new model predictions.
The bottleneck is usually the embedding of the input texts using the underlying embedding model.
These embeddings are cached to speed up future iterations.
Once the instances have been selected, they are shown in the UI.
The user now annotates all or a subset of the instances.
Optionally, they can reveal the annotation of the current model to ease the annotation work.
However, instance predictions are not shown by default to avoid biasing the annotator.

The user can now upload the instances (\verb!update!) to the platform which results in retraining the underlying few-shot model.
At this point, the user can continue to iterate on the model by requesting more instances.
When the user is satisfied with the current model they can call model inference on a set of instances (\verb!run!).

\subsection{Performance Prediction}
\label{sec:perf_pred}

A critical question is how the user knows when to stop annotating (\verb!evaluate!).
To ease their decision making, we add a number of metrics that can be computed even when no test instances is available, which is typically the case in FSL.
In particular, we implemented cross-validation on the labeled instances and various metrics that do not require any labels.
We collect a random sample $T$ of 1,000 unlabeled training instances.
After every AL iteration $i$ we assign the current model distribution $\mathrm{P}_i(Y \mid X)$ to these instances.
We then define metrics that are computed for every instance and averaged over the entire sample:

\begin{itemize}
\item \textbf{Negative Entropy}:\\ $-\mathrm{H}_{i}(Y) = \sum_{j} \mathrm{P}_i( y_j \mid x) \log \mathrm{P}_i( y_j \mid x)$
\item \textbf{Max Prob}:  $\mathrm{P}_i (\hat{y} \mid x)$, with $\hat{y}_i$ as the most probable class
\item \textbf{Margin}: $\left[ \mathrm{P}_i( \hat{y}_{i,1} \mid x) - \mathrm{P}_i(\hat{y}_{i,2} \mid x) \right]$, with $\hat{y}_{i,k}$ as kth most probable class
\item \textbf{Negative Update Rate}: $\delta(\hat{y}_{i-1}(x), \hat{y}_{i}(x))$, with $\delta$ as the Kronecker delta
\item \textbf{Negative Kullback-Leibler divergence}: $-\displaystyle D_{\text{KL}}(P_i\parallel P_{i-1})$
\end{itemize}

\emph{Entropy}, \emph{max prob} and \emph{margin} are based on the uncertainty measure used in AL.
The intuition is that the uncertainty of the model is reduced and converges as the model is trained.
The \emph{update rate} denotes the relative number of instances with a different model prediction as in the previous iteration.
Again we assume that this rate lowers and converges as the model training converges.
The \emph{KL divergence} provides a more sensitive version of the update rate that considers the entire label distribution.

\begin{figure*}[t]
\centering
\resizebox{\textwidth}{!}{%
\includegraphics[]{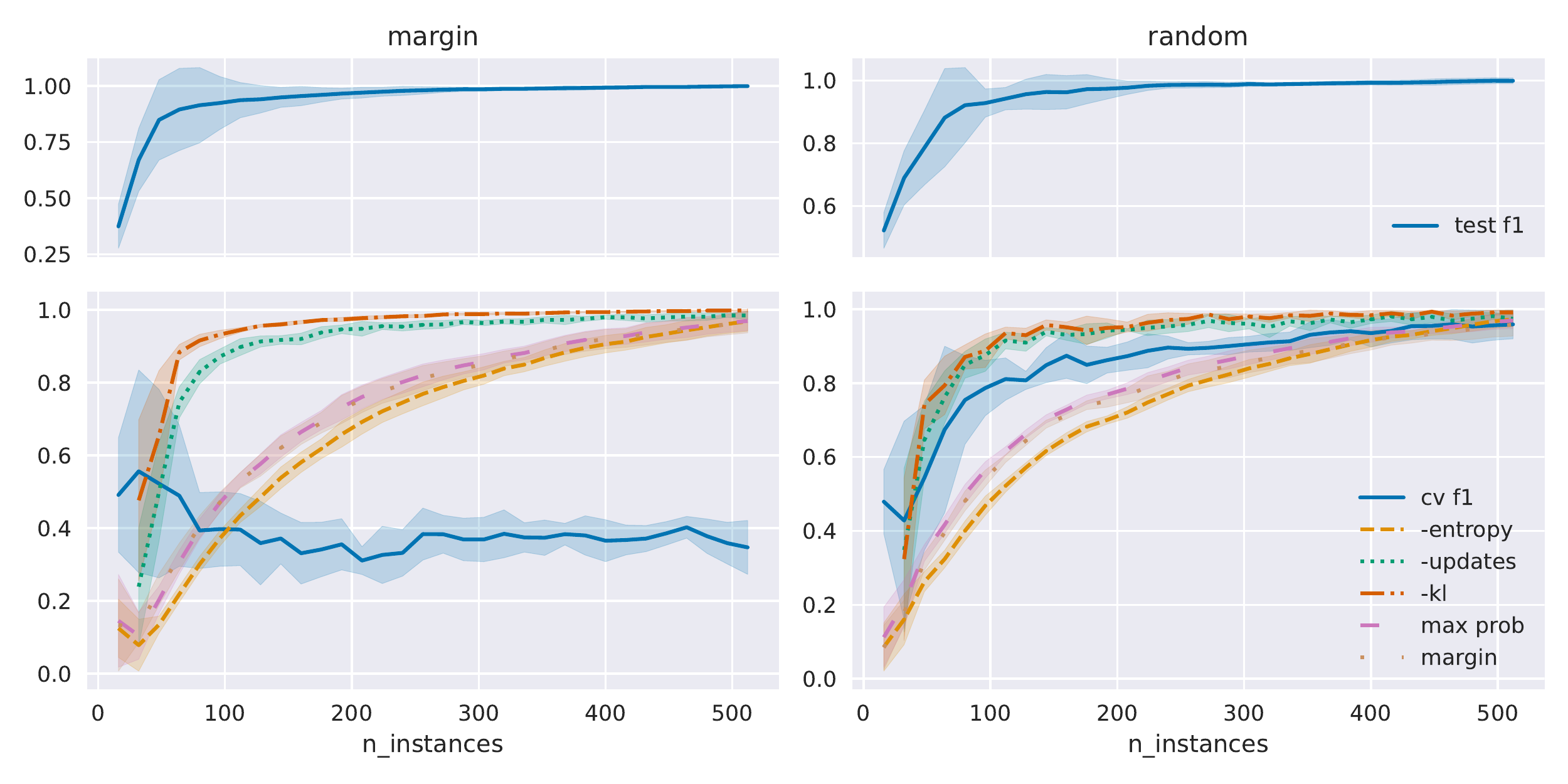}
}
\caption{Training metrics for \agnews for the \emph{random} and margin AL methods. The upper plot shows the true average normalized test F1 (blue curve) as well as the average error of the regressor model (blue shade).
The lower plot shows the raw training metrics.}
\label{fig:train_metrics}
\end{figure*}

We also experiment with combining these signals in a random forest regressor (RFR) \cite{ho1995random}.
For every iteration $i$ the model predicts the normalized test F1.
In general, it is hard to predict the true F1 for an unknown classification problem and dataset without even knowing the test set.
Therefore, we normalize all target test F1 curves by dividing by their maximum value (Figure \ref{fig:train_metrics}).
Intuitively, the RFR tells us how much of the performance that we will ever reach on this task we have reached so far.

The feature set of the model consists of a few base features such as the number of instances the model has been trained with, the AL method used and the number of labels.
Additionally, for each of the metrics above, we add the value at the current iteration $i$ as well as of a history of the last $h=5$ iterations.

\begin{figure*}[t]
\resizebox{\textwidth}{!}{%
\includegraphics[]{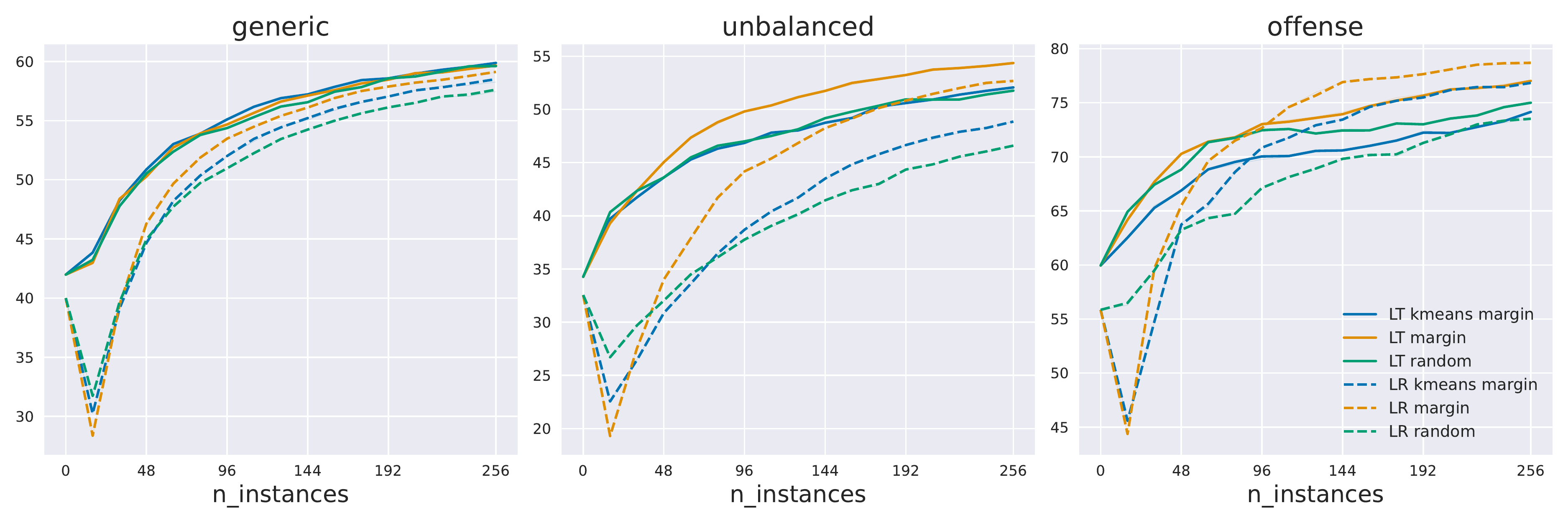}
}
\caption{Average test macro F1 score on the generic, unbalanced and offensive datasets. LT and LR denote label tuning and logistic regression, respectively.}
\label{fig:main_plot}
\end{figure*}

\begin{table*}[t]
\centering
\resizebox{0.7\textwidth}{!}{%


\begin{tabular}{l@{\hskip 0.25cm}l@{\hskip 0.25cm}l@{\hskip 0.25cm}l@{\hskip 0.25cm}l@{\hskip 0.25cm}l@{\hskip 0.25cm}l@{\hskip 0.25cm}l@{\hskip 0.25cm}l}
\toprule
\parbox[t]{2mm}{\multirow{11}{*}{\rotatebox[origin=c]{90}{Generic}}}
           &name &         gnad &              azn-de &            ag-news &              azn-en &              azn-es &                 hqa &        mean \\
\cmidrule{2-9}
           &LT random & 72.3$_{1.4}$ & \bftab 47.7$_{0.6}$ &        85.3$_{1.0}$ &        48.3$_{1.1}$ &        46.4$_{1.5}$ &        57.7$_{1.6}$ &        59.6 \\
\cmidrule{2-9}
   &LT kmeans margin & 74.6$_{1.1}$ &        47.5$_{1.6}$ &        86.0$_{0.4}$ &        46.7$_{1.4}$ &        45.9$_{1.7}$ &        58.5$_{0.9}$ & \bftab 59.9 \\
 &LT kmedoids margin & 73.8$_{1.5}$ &        45.9$_{1.9}$ &        86.0$_{0.5}$ &        48.3$_{1.3}$ &        46.3$_{1.3}$ &        57.8$_{1.4}$ &        59.7 \\
          & LT margin & \bftab 74.9$_{1.5}$ &        45.5$_{1.3}$ & \bftab 86.2$_{0.7}$ &        46.6$_{1.4}$ &        46.0$_{1.4}$ & \bftab 58.6$_{0.9}$ &        59.7 \\
        &LT kmedoids & 70.9$_{2.5}$ &        47.5$_{1.2}$ &        84.9$_{1.0}$ &        48.1$_{1.1}$ & \bftab 46.7$_{1.6}$ &        56.9$_{1.1}$ &        59.1 \\
         &LT k-means & 70.9$_{1.2}$ &        47.4$_{1.7}$ &        82.8$_{1.2}$ & \bftab 48.6$_{0.9}$ &        46.0$_{1.3}$ &        55.4$_{1.2}$ &        58.5 \\
             &LT CAL & 69.5$_{2.3}$ &        47.0$_{1.4}$ &        84.4$_{0.7}$ &        47.9$_{1.6}$ &        46.0$_{1.1}$ &        56.0$_{2.4}$ &        58.5 \\
&LT kmedoids entropy & 73.3$_{1.2}$ &        41.1$_{2.6}$ &        85.1$_{0.6}$ &        43.2$_{2.6}$ &        41.6$_{2.1}$ &        57.7$_{1.5}$ &        57.0 \\
  &LT kmedoids least & 72.4$_{1.3}$ &        42.6$_{2.0}$ &        85.2$_{0.5}$ &        43.9$_{1.8}$ &        40.3$_{2.6}$ &        57.5$_{1.3}$ &        57.0 \\
         &LT entropy & 72.1$_{2.1}$ &        40.2$_{2.3}$ &        84.7$_{0.7}$ &        41.3$_{2.5}$ &        39.9$_{2.2}$ &        56.5$_{1.9}$ &        55.8 \\
\toprule
\parbox[t]{2mm}{\multirow{10}{*}{\rotatebox[origin=c]{90}{Unbalanced}}}
          &LT random &        62.6$_{6.6}$ &        38.8$_{2.4}$ &        81.9$_{1.0}$ &        39.2$_{2.5}$ &        42.5$_{3.0}$ & 45.5$_{2.0}$ &        51.8 \\
\cmidrule{2-9}
          &LT margin & 70.7$_{2.2}$ &        40.5$_{2.4}$ & \bftab 83.7$_{0.8}$ &        41.6$_{2.9}$ &        41.8$_{2.4}$ & 47.9$_{1.5}$ & \bftab 54.4 \\
             &LT CAL &        65.2$_{3.4}$ & \bftab 41.1$_{2.9}$ &        82.0$_{1.1}$ & \bftab 43.8$_{1.5}$ & \bftab 42.9$_{2.2}$ & 48.4$_{1.8}$ &        53.9 \\
         &LT entropy & \bftab 70.8$_{1.7}$ &        40.5$_{1.9}$ &        82.5$_{1.3}$ &        38.5$_{3.1}$ &        39.7$_{1.4}$ & 47.6$_{2.7}$ &        53.2 \\
  &LT kmedoids least &        65.4$_{2.8}$ &        39.2$_{2.1}$ &        83.2$_{0.9}$ &        42.5$_{2.1}$ &        39.2$_{3.0}$ & 48.0$_{1.5}$ &        52.9 \\
&LT least confidence &        69.1$_{2.4}$ &        38.4$_{2.1}$ &        83.5$_{0.7}$ &        39.7$_{2.0}$ &        38.8$_{1.5}$ & 46.4$_{2.4}$ &        52.7 \\
 &LT kmedoids margin &        62.0$_{5.0}$ &        40.5$_{3.0}$ &        83.6$_{0.5}$ &        40.8$_{1.6}$ &        40.6$_{2.4}$ & 47.4$_{1.9}$ &        52.5 \\
&LT kmedoids entropy &        64.1$_{2.9}$ &        39.0$_{2.6}$ &        82.6$_{1.0}$ &        39.6$_{1.8}$ &        40.4$_{1.1}$ & 47.6$_{2.4}$ &        52.2 \\
   &LT kmeans margin &        66.8$_{5.1}$ &        37.4$_{2.8}$ &        82.4$_{1.3}$ &        39.9$_{2.0}$ &        38.8$_{1.7}$ & 47.0$_{1.5}$ &        52.1 \\
        &LT kmedoids &        59.5$_{6.8}$ &        37.6$_{3.3}$ &        81.0$_{1.6}$ &        41.3$_{1.7}$ &        41.6$_{2.4}$ & 44.8$_{1.6}$ &        51.0 \\
\toprule
\parbox[t]{2mm}{\multirow{11}{*}{\rotatebox[origin=c]{90}{Offense}}}
               &name &                hate &               solid &&&&&        mean \\
\cmidrule{2-9}
          &LR random	&        61.8$_{4.6}$ & 	   85.2$_{1.9}$ &&&&&        73.5 \\
\cmidrule{2-9}
          &LR margin & \bftab 69.1$_{1.8}$ & \bftab 88.3$_{0.6}$ &&&&& \bftab 78.7 \\
&LR least confidence &        68.9$_{2.2}$ &        87.8$_{0.5}$ &&&&&        78.3 \\
         &LR entropy &        68.8$_{2.2}$ &        87.8$_{0.6}$ &&&&&        78.3 \\
             &LR CAL &        68.3$_{2.6}$ &        87.2$_{0.9}$ &&&&&        77.7 \\
  &LR kmedoids least &        66.7$_{2.1}$ &        87.5$_{0.9}$ &&&&&        77.1 \\
 &LR kmedoids margin &        66.4$_{2.4}$ &        87.3$_{1.0}$ &&&&&        76.9 \\
   &LR kmeans margin &        66.3$_{1.9}$ &        87.4$_{0.9}$ &&&&&        76.8 \\
&LR kmedoids entropy &        66.1$_{2.2}$ &        87.6$_{0.7}$ &&&&&        76.8 \\
        &LR kmedoids &        64.8$_{3.7}$ &        86.8$_{0.9}$ &&&&&        75.8 \\
\bottomrule
\end{tabular}

}
\caption{Average test macro F1 score on the generic, unbalanced and offense datasets for 256 instances. LT and LR denote label tuning and logistic regression. The subscript denotes the standard deviation and bold font indicates the column-wise maximum.}
\label{tab:results}
\end{table*}

\begin{table*}[t]
\centering
\resizebox{0.7\textwidth}{!}{%
\begin{tabular}{l@{\hskip 0.25cm}r@{\hskip 0.5cm}r@{\hskip 0.5cm}r@{\hskip 0.25cm}r@{\hskip 0.25cm}r@{\hskip 0.5cm}c@{\hskip 0.25cm}r@{\hskip 0.25cm}c}
\toprule
                model &         MSE &         AUC &          \fmetric &           \pmetric &           \rmetric &    test F1 &        err &    instances \\
\midrule
        baseline  272 &      3483.2 &        86.6 &        80.4 &        81.2 & \bftab 86.8 &        94.3 &        2.0 &        272.0 \\
        baseline  288 &      3759.8 &        86.3 &        80.0 &        83.1 &        84.0 &        94.6 &        1.8 &        288.0 \\
        baseline  304 &      4038.8 &        85.9 &        79.4 &        85.1 &        81.1 & \bftab 95.2 &        1.4 &        304.0 \\
\midrule
                 base &        73.5 & \bftab 97.3 &        76.5 &        83.1 &        80.1 &        93.9 &        2.5 &        284.6 \\
\midrule
        forward cv-f1 &        56.5 &        96.9 &        80.7 &        84.0 &        83.8 &        94.5 &        1.8 & \bftab 271.8 \\
  forward -entropy &        70.8 &        96.4 &        79.1 &        84.5 &        80.9 &        94.3 &        2.1 &        283.5 \\
  forward -updates &        61.1 &        96.6 &        80.2 &        86.0 &        81.1 &        94.7 &        1.5 &        284.4 \\
       forward -kl &        77.1 &        96.6 &        75.0 &        84.3 &        76.0 &        94.3 &        2.3 &        295.1 \\
     forward max-prob &        59.9 &        96.9 &        80.4 &        86.0 &        80.4 &        95.0 &        1.4 &        288.1 \\
       forward margin &        59.4 &        96.9 &        79.2 & \bftab 86.7 &        78.3 &        95.0 &        1.4 &        294.2 \\
\midrule
                  all &        65.9 &        97.1 &        80.4 &        86.2 &        81.6 &        95.1 & \bftab 1.3 &        287.3 \\
              all h=0 & \bftab 56.0 &        97.1 &        79.7 &        84.7 &        81.3 &        94.3 &        2.0 &        277.1 \\
              all h=1 &        57.4 &        97.1 &        80.7 &        85.4 &        82.4 &        94.5 &        1.7 &        276.6 \\
\midrule
       backward cv-f1 &        66.7 &        96.7 &        80.6 &        84.7 &        82.8 &        94.7 &        1.7 &        280.2 \\
 backward -entropy &        57.0 &        97.0 &        80.7 &        86.1 &        81.9 &        95.0 &        1.3 &        285.5 \\
 backward -updates &        64.5 &        97.1 &        80.4 &        86.3 &        81.1 &        95.0 &        1.4 &        288.6 \\
      backward -kl &        68.0 & \bftab 97.2 & \bftab 81.2 & \bftab 86.7 &        82.4 & \bftab 95.2 & \bftab 1.2 &        287.9 \\
    backward max-prob &        65.7 &        97.1 &        80.3 &        86.5 &        81.2 &        95.0 &        1.3 &        288.6 \\
      backward margin &        63.8 &        97.1 &        80.2 &        86.1 &        81.0 &        95.0 &        1.4 &        289.8 \\
\bottomrule
\end{tabular}

}
\caption{Results on the prediction of the test F1: Mean squared error (MSE in \textpertenthousand ), Area under curve (AUC), precision (\pmetric), recall (\rmetric) and \fmetric as well as average test F1, error and number of training instances reached at a threshold of $\tau=0.95$. 
\emph{baseline i} indicates a baseline that predict 0 for every step $j: j < i$ and 1 otherwise. \emph{base} is the base feature set discussed in S. \ref{sec:perf_pred}. \emph{all} uses all features. \emph{forward} and \emph{backward} denote ablation experiments with forward selection from \emph{base} and backward selection from \emph{all}, respectively. $h$ indicates the size of the history and is 5 by default.}
\label{tab:f1_prediction}
\end{table*}

\section{Related Work}

\paragraph{Active Learning Methods}
We evaluate AL methods that have been reported to give strong results in the literature.
Within \textit{uncertainty sampling}, \emph{margin sampling} has been found to out-perform other methods also for modern model architectures \cite{schroeder2021uncertaintybased,lu2020investigating}.
Regarding methods that combine  \textit{uncertainty} and \textit{diversity sampling}, CAL \cite{margatina-etal-2021-active} has been reported to give consistently better results than
BADGE \cite{Ash2020DeepBA} and ALPS \cite{Yuan2020ColdstartAL} on a range of datasets.
A line of research that we exclude are Bayesian approaches such as BALD \cite{houlsby2011bayesian}, because the requirement of a model ensemble makes them computationally inefficient.

\paragraph{Few-Shot Learning with Label Tuning}
Our work differs from much of the related work in that we use a particular model and training regimen: label tuning (LT) \cite{labeltuning}.
LT is an approach that only tunes a relative small set of parameters, while the underlying Siamese Network model \cite{reimers-2019-sentence-bert} remains unchanged.
This makes training fast and deployment scalable. More details are given in Section \ref{sec:al_models}.

\paragraph{Cold Start}
Cold start refers to the zero-shot case where we start without any training examples.
Most studies \cite{Ash2020DeepBA,margatina-etal-2021-active} do not work in this setup and start with a seed set of 100 to several thousand labeled examples.
\citet{griesshaber-etal-2020-fine,schroeder2021uncertaintybased,lu2020investigating} use few-shot ranges of less than thousand training examples but still use a seed set.
\citet{Yuan2020ColdstartAL} approach a zero-shot setting without an initial seed set but they differs from our work in model architecture and training regimen.

\paragraph{Balanced and Unbalanced Datasets}
Some studies have pointed out inconsistencies on how AL algorithms behave across different models or datasets \cite{Lowell2019PracticalOT}.
It is further known in the scientific community that AL often does not out-perform random selection when the label distribution is balanced.\footnote{\url{https://tinyurl.com/fasl-community}}
\citet{ein-dor-etal-2020-active} focus on unbalanced datasets but not with a cold start scenario.
To fill this gap, we run a large study on AL in the under-researched few-shot-with-cold-start scenario, looking into both balanced and unbalanced datasets.

\section{Experimental Setup}

We compare a number of active learning models on a wide range of datasets.

\begin{table*}[t]
\centering
\resizebox{0.5\textwidth}{!}{%
\begin{tabular}{l@{\hskip 0.25cm}l@{\hskip 0.25cm}r@{\hskip 0.25cm}r@{\hskip 0.25cm}c@{\hskip 0.25cm}r}
\toprule
        &dataset &  train &  test &  $|L|$ &  $U_\%$\\
\midrule
\multirow{6}{*}{Generic}
&\gnad \cite{Gnad} &   9,245 &  1,028 &       9 & 34.4 \\
&\agnews \cite{AgNews} & 120,000 &  7,600 &       4 &  0.0 \\
&\headqa \cite{headqa} &   4,023 &  2,742 &       6 &  2.8 \\
&\amazonreviews-de \cite{amazonreviews} & 205,000 &  5,000 &       5 &  0.0 \\
&\amazonreviews-en & 205,000 &  5,000 &       5 &  0.0 \\
&\amazonreviews-es & 205,000 &  5,000 &       5 &  0.0 \\
\midrule
\multirow{6}{*}{Unbalanced}
&\gnad        &   3,307 &   370 &       9 & 92.9 \\
&\agnews      &  56,250 &  3,563 &       4 & 60.0 \\
&\headqa      &   1,373 &   927 &       6 & 82.9 \\
&\amazonreviews-de             &  79,438 &  1,938 &       5 & 74.8 \\
&\amazonreviews-en             &  79,438 &  1,938 &       5 & 74.8 \\
&\amazonreviews-es &  79,438 &  1,938 &       5 & 74.8 \\
\midrule
\multirow{2}{*}{Offense}
& \hatespeech \cite{hatespeech18} &   8,703 &  2,000 &       2 & 77.8 \\
& \solid \cite{offenseval2020} &   1,887 &  2,000 &       2 & 45.6 \\
\bottomrule
\end{tabular}
}
\caption{
Dataset statistics. train and test sizes of the splits. $|L|$ is the cardinality of the label set $L$. $\mathrm{U}$ quantifies the uniformness: $\sum_{l \in L} \left|f(l) - \frac{1}{|L|} \right|$, where $f(l)$ is the relative frequency of label $l$. $U=0$ indicates that the data is distributed uniformly. Note that the generic datasets are balanced while the others are skewed.
}
\label{tab:datastats}
\end{table*}

\subsection{Datasets}

\paragraph{Generic Datasets} We run experiments on 4 generic text classification datasets in 3 different languages.
AG News (\agnews) \cite{AgNews} and GNAD (\gnad) \cite{Gnad} are news topic classification tasks in English and German, respectively.
Head QA (\headqa) \cite{headqa} is a Spanish catalogue of questions in a health domain that are grouped into categories such as medicine, biology and pharmacy.
Amazon Reviews (\amazonreviews) \cite{amazonreviews} is a large corpus of product reviews with a 5-star rating in multiple languages. Here we use the English, Spanish and German portion.

\paragraph{Unbalanced Datasets} All of these datasets have a relatively even label distribution. To investigate the performance of AL in a more difficult setup, we also create a version of each dataset where we
enforce a label distribution with exponential decay. We implement this by down-sampling some of the labels, without replacement. In particular, we use the adjusted frequency $n'(y) = n(\hat{y})^{-2 \cdot \operatorname{rank}(y)}$, where $n(y)$ is the original frequency of label $y$, $\hat{y}$ is the most frequent label and $\operatorname{rank}(y)$ is the frequency rank with $\operatorname{rank}(\hat{y}) = 0$.

\paragraph{Offensive Language Datasets}
We also evaluate on the Semi-Supervised Offensive Language Identification Dataset (\solid) \cite{offenseval2020} and HateSpeech 2018 (\hatespeech) \cite{hatespeech18}.
These dataset are naturally unbalanced and thus suited for AL experiments.

Table \ref{tab:datastats} provides statistics on the individual datasets.
Note that -- following other work in few-shot learning \cite{Wang2021EntailmentAF,schick-schutze-2021-exploiting} -- we do not use a validation set.
This is because in a real world FSL setup one would also not have access to any kind of evaluation data.
As a consequence, we do not tune hyper-parameters in any way and use the defaults of the respective frameworks.
The label descriptions we use are taken from the related work \cite{Wang2021EntailmentAF,labeltuning} and can be found in the Appendix (S. \ref{sec:appendix}).

\subsection{Simulated User Experiments}
It is common practice \cite{Lowell2019PracticalOT} in AL research to simulate the annotation process using labeled datasets.
For every batch of selected instances, the gold labels are revealed, simulating the labeling process of a human annotator.
Naturally, a simulation is not equivalent to real user studies and certain aspects of the model cannot be evaluated.
For example, some methods might retrieve harder or more ambiguous examples that will result in more costly annotation and a higher error rate.
Still we chose simulation in our experiments as they are a scalable and reproducible way to compare the quality of a large number of different methods.
In all experiments, we start with a zero-shot model that has not been trained on any instances.
This sets this work apart from most related work that starts with a model trained on a large seed set of usually thousands of examples.
We then iteratively select batches of $k=16$ instances until we reach a training set size of 256.
The instances are selected from the entire training set of the respective dataset. However, to reduce the computational cost we down-sample each training set to at most 20,000 examples.

As FSL with few instances is prone to yield high variance on the test predictions we average all experiments over 10 random trials.
We also increase the randomness of the instance selection by first selecting $2k$ instances with the respective method and later sampling $k$ random examples from the initial selection.

\section{Results}
\label{sec:results}

\paragraph{Active Learning Methods}
Figure \ref{fig:main_plot} shows the AL progression averaged over multiple datasets.
For improved clarity only the best performing methods are shown.
Plots for the individual dataset can be found in the Appendix (S. \ref{sec:appendix}).
On the comparison between Label Tuning (LT) and Logistic Regression (LR) we find that LT outperforms LR on the generic datasets as well as the unbalanced datasets.
On the offense datasets the results are mixed: LR outperforms LT when using \emph{margin}, but is outperformed when using \emph{random}.

With respect to the best performing selection methods we find that \emph{random} and \emph{margin} perform equally well on the generic datasets when LT is used.
However, when using LR or facing unbalanced label distributions, we see that \emph{margin} gives better results than \emph{random}.
On the offense datasets we also find substantial differences between \emph{margin} and \emph{random}. 

Regarding \emph{kmeans-margin}, we see mixed results. In general, \emph{kmeans-margin} does not outperform \emph{random} when LT is used but does so when LR is used.
In most experiments \emph{kmeans-margin} is inferior to \emph{margin}.

Table \ref{tab:results} shows results by dataset for the 10 best performing methods. For each set of datasets we include the 10 best performing methods of the best performing model type.
We find that uncertainty sampling (such as \emph{margin} and \emph{least-confidence}) methods outperform diversity sampling methods (such as \emph{kmeans} and \emph{kmedoids}).
Hybrid methods such as (\emph{CAL} and \emph{kmeans-margin}) perform better than pure diversity sampling methods but are in general behind their counterparts based on importance sampling.

\paragraph{Test F1 Prediction Experiments}
We test the F1 regression models with the \emph{margin} and \emph{random} AL method.
We trained the LR model with $k=16$ instances per iteration until reaching a training set size of 512.
As for the AL experiments above we repeat every training run 10 times.
We then train a prediction model for each of the 12 datasets used in this study (leaving out one dataset at a time).

Figure \ref{fig:train_metrics} shows the model trained for \agnews.
We see that the cross-validation F1 computed on the labeled instances (cv f1) has a similar trend as the test F1 when we use random selection.
However, when using AL with margin selection the curve changes drastically.
This indicates that CV is not a good evaluation method or stopping criterion when using AL.
The uncertainty based metrics (max prob, margin and -entropy) correlate better with test F1 but still behave differently in terms of convergence.
Negative KL divergence (-kl) and update rate (-updates) on the other hand show a stronger correlation.
With regard to the regressor model we can see that the average error (shaded blue area) is large in the beginning but drops to a negligible amount as we reach the center of the training curve.

For evaluating the model, we compute a range of metrics (Table \ref{tab:f1_prediction}).
Mean squared error (MSE) is a standard regression metric but not suited to the problem as it overemphasizes the errors at the beginning of the curve.
Therefore we define $\tau=0.95$ as the threshold that we are most interested in.
That is, we assume that the user wants to train the model until reaching 95\% of the possible test F1.
With $\tau$ we can define a binary classification task and compute AUC, recall (\rmetric), precision (\pmetric) and \fmetric.\footnote{Note \fmetric is the F1-score on the classification problem of predicting if the true test-F1 is $> \tau$, while test-F1 is the actual F1 score reached by the FSL model on the text classification task.}
Additionally, we compute the average test-F1, error (err) and training set size when the model first predicts a value $>\tau$.

In the ablation study with forward selection, we find that all features except \emph{-entropy} and \emph{-kl} lower the error rate (err) compared to \emph{base}.
For backward selection, only removing \emph{cv f1} causes a bigger increase in error rate compared to \emph{all}.
This might be because the unsupervised metrics are strongly correlated.
Finally, reducing the history (S. \ref{sec:perf_pred}) also causes an increase in error rate.

In comparison with the simple baselines (baseline $i$), where we simply stop after a fixed number of instances $i$, we find that \emph{all} gives higher \emph{test-f1} (95.1 vs 94.6) at comparable instance numbers (287 vs 288).
This indicates that a regression model adds value for the user.

\section{Conclusion}

We studied the problem of active learning in a few-shot setting. We found that margin selection outperforms random selection for most models and setups, unless the labels of the task are distributed uniformly.
We also looked into the problem of performance prediction to compensate the missing validation set in FSL.
We showed that the normalized F1 on unseen test data can be approximated with a random forest regressor (RFR) using signals computed on unlabeled instances.
In particular, we showed that the RFR peforms better than the baseline of stopping after a fixed number of steps.
Our findings have been integrated into \fasl, a uniform platform with a UI that allows non-experts to create text classification models with little effort and expertise.

\section{Appendix}
\label{sec:appendix}

The repository at \url{https://tinyurl.com/fasl-gh} contains the label descriptions,  UI screenshots and additional plots and results for active learning and performance prediction.

\section*{Acknowledgements}

We gracefully thank the support of the Pro$^2$Haters - Proactive Profiling of Hate Speech Spreaders (CDTi IDI-20210776), DETEMP - Early Detection of Depression Detection in Social Media (IVACE IMINOD/2021/72) and DeepPattern (PROMETEO/2019/121) R\&D grants. Grant PLEC2021-007681 funded by MCIN/AEI/ 10.13039/501100011033 and by European Union NextGenerationEU/PRTR.


\bibliography{references}
\end{document}